\pdfoutput=1

\documentclass[11pt]{article}

\usepackage{times}
\usepackage{latexsym}

\usepackage[T1]{fontenc}

\usepackage[utf8]{inputenc}

\usepackage{microtype}

\usepackage{microtype}
\usepackage{graphicx}
\usepackage{subfigure}
\usepackage{booktabs} 
\usepackage{natbib}
\usepackage{graphicx}
\usepackage{amsmath}
\usepackage{amscd}
\usepackage{amssymb}
\usepackage{amsthm}
\usepackage{mathabx}
\usepackage{algorithmic}
\usepackage{algorithm}
\usepackage{booktabs}

\usepackage{multirow}
\usepackage{paralist}
\usepackage{color, colortbl}
\usepackage{enumitem}
\usepackage[normalem]{ulem}
\usepackage{stackengine}

\usepackage{xcolor}
\usepackage{tikz}
\usetikzlibrary{automata, arrows}
\usepackage{pgfplots}

\usetikzlibrary{arrows, positioning, calc}
\definecolor{color1}{RGB}{27,158,119}
\definecolor{color3}{RGB}{217,95,2}
\definecolor{color2}{RGB}{117,112,179}

\usetikzlibrary{shapes.geometric}
\newcommand{\En}{$\texttt{En}$}
\newcommand{\Zh}{$\texttt{Zh}$}
\newcommand{\Gu}{$\texttt{Gu}$}
\newcommand{\Hi}{$\texttt{Hi}$}
\newcommand{\Kk}{$\texttt{Kk}$}
\newcommand{\Ru}{$\texttt{Ru}$}
\newcommand{\Tr}{$\texttt{Tr}$}
\newcommand{\Ar}{$\texttt{Ar}$}
\newcommand{\Ta}{$\texttt{Ta}$}
\newcommand{\Te}{$\texttt{Te}$}
\newcommand{\Ne}{$\texttt{Ne}$}
\newcommand{\Si}{$\texttt{Si}$}
\newcommand{\X}{$\texttt{X}$}
\newcommand{\Y}{$\texttt{Y}$}

\usepackage{naacl2021}

\usepackage{times}
\usepackage{latexsym}

\usepackage[T1]{fontenc}

\usepackage[utf8]{inputenc}

\usepackage{microtype}

\title{Harnessing Multilinguality in Unsupervised Machine Translation for Rare Languages}
\author{Xavier Garcia \and Aditya Siddhant \and Orhan Firat \and Ankur P. Parikh \\
         Google Research \\
         Mountain View \\
         California \\
         \texttt{{xgarcia,adisid,orhanf,aparikh}@google.com}
         }
\begin{document}
\pgfplotsset{compat=1.6}
\maketitle
\begin{abstract}

Unsupervised translation has reached impressive performance on resource-rich language pairs such as English-French and English-German. However, early studies have shown that in more realistic settings involving low-resource, rare languages, unsupervised translation performs poorly, achieving less than 3.0 BLEU. In this work, we show that \textit{multilinguality} is critical to making unsupervised systems practical for low-resource settings. In particular, we present a single model for 5 low-resource languages (Gujarati, Kazakh, Nepali, Sinhala, and Turkish) to and from English directions, which leverages monolingual and auxiliary parallel data from other high-resource language pairs via a three-stage training scheme. We outperform all current state-of-the-art unsupervised baselines for these languages, achieving gains of up to 14.4 BLEU. Additionally, we outperform strong supervised baselines for various language pairs as well as match the performance of the current state-of-the-art supervised model for \Ne $\rightarrow$\En . We conduct a series of ablation studies to establish the robustness of our model under different degrees of data quality, as well as to analyze the factors which led to the superior performance of the proposed approach over traditional unsupervised models.

\end{abstract}

\section{Introduction}

Neural machine translation systems \cite{kalchbrenner13,sutskever14,bahdanau14,wu2016google} have demonstrated state-of-the-art results for a diverse set of language pairs when given large amounts of relevant parallel data. However, given the prohibitive nature of such a requirement for low-resource language pairs, there has been a growing interest in unsupervised machine translation~\cite{ravi11} and its neural counterpart, unsupervised neural machine translation (UNMT)~\cite{lample17,artetxe17}, which leverage only monolingual source and target corpora for learning. Bilingual unsupervised systems~\citep{lample19,artetxe19,ren19,D2GPO} have achieved surprisingly strong results on high-resource language pairs such as English-French and English-German.

However, these works only evaluate  on high-resource language pairs with high-quality data, which are not realistic scenarios where UNMT would be utilized. Rather, the practical potential of UNMT is in low-resource, rare languages that may not only lack parallel data but also have a shortage of high-quality monolingual data. For instance, Romanian (a typical evaluation language for  unsupervised methods) has 21 million lines of high-quality in-domain monolingual data provided by WMT. In contrast, for an actual low-resource language, Gujarati, WMT only provides 500 thousand lines of monolingual data (in news domain) and an additional 3.7 million lines of monolingual data from Common Crawl (noisy, general-domain).

Given the comparably sterile setups UNMT has been studied in, recent works have questioned the usefulness of UNMT when applied to more realistic low-resource settings. \citet{kim2020and} report BLEU scores of less than 3.0 on low-resource pairs and \citet{marchisio2020does} also report dramatic degradation under domain shift.

However, the negative results shown by the work above only study bilingual unsupervised systems and do not consider \textit{multilinguality}, which has been well explored in supervised, zero-resource and zero-shot settings~\citep{johnson16,firat16,firat-etal-2016-zero,DBLP:journals/corr/ChenLCL17,neubig2018rapid,gu2018universal,liu2020multilingual,ren2018triangular, zoph16} to improve performance for low-resource languages. The goal of this work is to study if multilinguality can help UNMT be more robust in the low-resource, rare language setting.

\begin{figure}[t!]
  \centering
  \def \radius {2.3cm}
  \begin{tikzpicture}[>=stealth',shorten >=1pt,auto,node distance=1.7cm]
    \node[state]    at (0,0)     (en) [] {$\textbf{En}$};
    \node[state]    at ({360/11 * (1 - 1)}:\radius)     (si) [] {$\textbf{Si}$};
    \node[state]    at ({360/11 * (2 - 1)}:\radius)     (zh)  {$\textbf{Zh}$};
    \node[state]    at ({360/11 * (3 - 1)}:\radius)     (tr)  {$\textbf{Tr}$};
    \node[state]    at ({360/11 * (4 - 1)}:\radius)     (ta)   {$\textbf{Ta}$};
    \node[state]    at ({360/11 * (5 - 1)}:\radius)     (te)  {$\textbf{Te}$};
    \node[state]    at ({360/11 * (6 - 1)}:\radius)    (ne)  {$\textbf{Ne}$};
    \node[state]    at ({360/11 * (7 - 1)}:\radius)    (hi) {$\textbf{Hi}$};
    \node[state]    at ({360/11 * (8 - 1)}:\radius)     (kk)  {$\textbf{Kk}$};
    \node[state]    at ({360/11 * (9 - 1)}:\radius)     (ar)  {$\textbf{Ar}$};
    \node[state]    at ({360/11 * (10 - 1)}:\radius)     (gu)  {$\textbf{Gu}$};
    \node[state]    at ({360/11 * (11 - 1)}:\radius)     (ru)   {$\textbf{Ru}$};

    \path[-]          (en)  edge                 node {} (hi);
    \path[-]          (en)  edge                 node {} (zh);
    \path[<->, color=blue, thick]          (en)  edge      [ dotted]           node {} (gu);
    \path[<->, color=blue, thick]          (en)  edge      [ dotted]           node {} (tr);
    \path[<->, color=blue, thick]          (en)  edge      [ dotted]           node {} (ne);
    \path[<->, color=blue, thick]          (en)  edge      [ dotted]           node {} (si);
    \path[<->, color=blue, thick]          (en)  edge      [ dotted]           node {} (kk);
    \path[-]          (en)  edge                 node {} (ta);
    \path[-]          (en)  edge                 node {} (ar);
    \path[-]          (en)  edge                 node {} (ru);
    \path[-]          (en)  edge                 node {} (te);

  \end{tikzpicture}
  \caption{A pictorial depiction of our setup. The dashed edge indicates the target unsupervised language pairs that lack parallel training data. Full edges indicate the existence of parallel training data. }
  \label{fig:langgraph}
\end{figure}
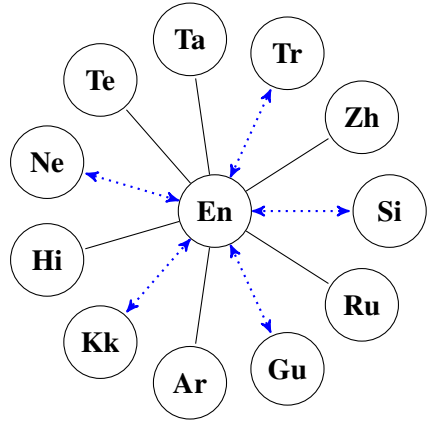

In our setup (Figure~\ref{fig:langgraph}), we have a single model for 5 target low-resource unsupervised directions (that are not associated with any parallel data): Gujarati, Kazakh, Nepali, Sinhala, and Turkish. These languages are chosen to be studied for a variety of reasons (discussed in \S\ref{sec:languages}) and have been of particular challenge to unsupervised systems. In our approach, as shown in Figure~\ref{fig:langgraph}, we also leverage auxiliary data from a set of higher resource languages: Russian, Chinese, Hindi, Arabic, Tamil, and Telugu. These higher resource languages not only possess significant amounts of monolingual data but also auxiliary parallel data with English that we leverage to improve the performance of the target unsupervised directions\footnote{ This makes our setting considerably more challenging than the zero-shot/zero resource setting. See \S\ref{sec:related} and \S\ref{sec:terminology} for a discussion.}.

Existing work on multilingual unsupervised translation~\citep{liu2020multilingual, garcia20,li2020reference,bai2020unsupervised}, which also uses auxiliary parallel data, employs a two-stage training scheme consisting of pre-training with noisy reconstruction objectives and fine-tuning with on-the-fly (iterative) back-translation and cross-translation terms (\S\ref{sec:background}). We show this leads to sub-optimal performance for low-resource pairs and propose an additional intermediate training stage in our approach. Our key insight is that pre-training typically results in high \X $\rightarrow$\En \ (to English) performance but poor \En $\rightarrow$\X \ (from English) results, which makes fine-tuning unstable. Thus, after pre-training, we propose an intermediate training stage that leverages offline back-translation~\citep{sennrich16} to generate synthetic data from the \X $\rightarrow$\En \, direction to boost \En $\rightarrow$\X \ accuracy. 

Our final results show that our approach outperforms a variety of supervised and unsupervised baselines, including the current state-of-the-art supervised model for the \Ne $\rightarrow$\En \, language pair. Additionally, we perform a series of experimental studies to analyze the factors that affect the performance of the proposed approach, as well as the performance in data-starved settings and settings where we only have access to noisy, multi-domain monolingual data.

\section{Related work}
\label{sec:related}

\begin{table*}[!htbp]
\tiny
\begin{tabular}{llcccccccccccc}
\toprule
 & \textbf{Domain} & \multicolumn{1}{l}{\, \En} & \multicolumn{1}{l}{\, \stackanchor{\Tr}{News}} & \multicolumn{1}{l}{\, \stackanchor{\Kk}{News}} & \multicolumn{1}{l}{\, \stackanchor{\Gu}{News}} & \multicolumn{1}{l}{\, \stackanchor{\Ne}{Wiki}} & \multicolumn{1}{l}{\, \stackanchor{\Si}{Wiki}} & \multicolumn{1}{l}{\, \stackanchor{\Te}{Wiki}} & \multicolumn{1}{l}{\, \stackanchor{\Ta}{Wiki}} & \multicolumn{1}{l}{\, \stackanchor{\Hi}{IITB}} & \multicolumn{1}{l}{\, \stackanchor{\Ru}{UN}} & \multicolumn{1}{l}{\, \stackanchor{\Ar}{UN}} & \multicolumn{1}{l}{\, \stackanchor{\Zh}{UN}}  \\ \midrule
\multirow{3}{*}{Monolingual} & News & 233M & 17M & 1.8M & 530K & - & - & 2.5M & 2.3M & 32.6M & 93.8M & 9.2M & 4.7M \\ 
& Wikipedia & - & - & - & 384K & 92k & 155k & - & - & - & - & - & 22.7M \\ 
& Crawled & - & - & 7.1M & 3.7M & 3.5M & 5.1M & - & - & - & - & - & - \\ \hline
\multirow{1}{*}{Auxiliary parallel (w/ English)} & Mixed & - & 205k & 225k & 10k & 564k & 647k & 290K & 350K & 1.5M & 23.2M & 9.2M & 15.8M \\ \hline
 \multirow{1}{*}{In-domain (\%)} & \quad - & - & 100\% & 20.2\% & 11.4\% & 2.0\% & 2.9\% & - & - & - & - & - & - \\ \bottomrule
\end{tabular}
\caption{The amount and domain of the data used in these experiments. For the unsupervised language pairs, we additionally included the domain of the development and test sets. For Arabic, we took the 18.4M samples from the UN Corpus and divided it in two, treating one half of it as unpaired monolingual data.  We include the amount of parallel data for the unsupervised language pairs, which is only utilized for our in-house supervised baseline.}
\label{tab:languages}
\end{table*}

Multilinguality has been extensively studied in the supervised literature and has been applied to the related problem of zero-shot translation~\citep{johnson16,firat16,arivazhagan2019missing,al19}.
Zero-shot translation concerns the case where direct (source, target) parallel data is lacking but there is parallel data via a common pivot language to both the source and the target. For example, in Figure~\ref{fig:langgraph}, \Ru $\leftrightarrow$\Zh \, and \Hi $\leftrightarrow$\Te \, would be zero-shot directions.

In contrast, a defining characteristic of the multilingual UNMT setup is that the source and target are disconnected in the graph and  one of the languages is not associated with any parallel data with English or otherwise.  \En $\leftrightarrow$\Gu \, or \En $\leftrightarrow$\Kk \ are such example pairs as shown in Figure~\ref{fig:langgraph}. 


Recently~\citet{guzman2019flores,liu2020multilingual} showed some initial results on multilingual unsupervised translation in the low-resource setting. They tune language-specific models and employ a standard two-stage training scheme~\citep{lample19}, or in the case of~\citet{liu2020multilingual} directly fine-tuning on a related language pair (e.g. \Hi $\rightarrow$\En) and then test on the target \X $\rightarrow$\En \ pair (e.g. \Gu $\rightarrow$\En). In contrast our approach trains one model for all the language pairs targetted and employs a three stage training scheme that leverages synthetic parallel data via offline back-translation.

Offline backtranslation~\citep{sennrich16} was originally used for unsupervised translation~\citep{lample18,artetxe19}, especially with phrase-based systems.

\subsection{Terminology}\label{subsec:terminology}
\label{sec:terminology}
There is some disagreement on the definition of multilingual unsupervised machine translation, which we believe arises from extrapolating unsupervised translation to multiple languages. In the case of only two languages, the definition is clear: unsupervised machine translation consists of the case where there is no parallel data between the source and target languages. However, in a setting with multiple languages, there are multiple scenarios which satisfy this condition. More explicitly, suppose that we want to translate between languages $\mathcal{X}$ and $\mathcal{Y}$ and we have access to data from another language $\mathcal{Z}$. Then, we have three possible scenarios:
\begin{itemize}
    \item We possess parallel data for $(\mathcal{X},\mathcal{Z})$ and $(\mathcal{Z},\mathcal{Y})$ which would permit a 2-step supervised baseline via the pivot. Existing literature \cite{johnson16,firat-etal-2016-zero} has used the term ``zero-shot" and ``zero-resource" to refer specifically to this setup. 
    \item We have parallel data for $(\mathcal{X},\mathcal{Z})$ but only monolingual data in $\mathcal{Y}$, as considered in \cite{li2020reference,liu2020multilingual,garcia20,bai2020unsupervised,guzman2019flores,artetxe2020call}. Note that the pivot-based baseline above is not possible in this setup.
    \item We do not have any parallel data among any of the language pairs, as considered in \cite{liu2020multilingual,sun2020knowledge}.
\end{itemize}


We believe the first setting is not particularly suited for the case where either $\mathcal{X}$ or $\mathcal{Y}$ are true low-resource languages (or extremely low-resource languages), since it is unlikely that these languages possess any parallel data with any other language. On the other hand, we usually assume that one of these languages is English and we can commonly find large amounts of parallel data for English with other high-resource auxiliary languages. For these reasons, we focus on the second setting for the rest of this work.

Arguably, the existence of the auxiliary parallel data provides some notion of indirect supervision that is not present when only utilizing monolingual data. However, this signal is weaker than the one encountered in the zero-shot setting, since it precludes the 2-step supervised baseline. As a result, recent work \cite{artetxe2020call,guzman2019flores,garcia20, liu2020multilingual} has also opted to use the term ``unsupervised". We too follow this convention and use this terminology, but we emphasize that independent of notation, our goal is to study the setting where \textit{only} the (extremely) low-resource languages of interest possess \textit{no} parallel data, whether with English or otherwise.

\section{Choice of languages}

\label{sec:languages}
The vast majority of works in UNMT (multilingual or otherwise) have focused on traditionally high-resource languages, such as French and German. While certain works simulate this setting by using only a smaller subset of the available monolingual data, such settings neglect common properties of true low-resource, rare languages: little-to-no lexical overlap with English and noisy data sources coming from multiple domains. Given the multi-faceted nature of what it means to be a low-resource language, we have chosen a set of languages with many of these characteristics.  We give a detailed account of the available data in Table \ref{tab:languages}.

\paragraph{Target unsupervised directions:} We select Turkish (\Tr), Gujarati (\Gu), and Kazakh (\Kk) from WMT . The latter two possess much smaller amounts of data than most language pairs considered for UNMT e.g. French or German. In order to vary the domain of our test sets, we additionally include Nepali (\Ne) and Sinhala (\Si) from the recently-introduced FLoRes dataset \cite{guzman2019flores}, as the test sets for these languages are drawn from Wikipedia instead of news. Not only do these languages possess monolingual data in amounts comparable to the low-resource languages from WMT, the subset of in-domain monolingual data for both languages make up less than 5\% of the available monolingual data of each language.
\paragraph{Auxiliary languages:} To choose our auxiliary languages that contain both monolingual data and parallel data with English, we took into account linguistic diversity, size, and relatedness to the target directions. Russian shares the same alphabet with Kazakh, and Hindi, Telugu, and Tamil are related to Gujarai, Nepali and Sinhala. Chinese, while not specifically related to any of the target language, is high resource and considerably different in structure from the other languages.
\section{Background}
\label{sec:background}
For a given language pair $(\texttt{X}, \texttt{Y})$ of languages $\texttt{X}$ and $\texttt{Y}$, we possess \emph{monolingual} datasets $\mathcal{D}_{\texttt{X}}$ and $\mathcal{D}_{\texttt{Y}}$, consisting of unpaired sentences of each language. 

\paragraph{Neural machine translation} In supervised neural machine translation, we have access to a \emph{parallel} dataset $\mathcal{D}_{\texttt{X} \times \texttt{Z}}$ consisting of translation pairs $(x,z)$. We then train a model by utilizing the \emph{cross-entropy} objective: $$\mathcal{L}_{\text{cross-entropy}}(x,y) = - \log p_{\theta}(y|x)$$ where $p_{\theta}$ is our translation model. We further assume $p_{\theta}$ follows the encoder-decoder paradigm, where there exists an encoder $\text{Enc}_{\theta}$ which converts $x$ into a variable-length representation which is passed to a decoder $p_{\theta}(y|x) := p_{\theta}(y| \text{Enc}_{\theta}(x))$.

\paragraph{Unsupervised machine translation} In this setup, we no longer possess $\mathcal{D}_{\texttt{X} \times \texttt{Y}}$. Nevertheless, we may possess auxiliary parallel datasets such as $\mathcal{D}_{\texttt{X} \times \texttt{Z}}$ for some language $\texttt{Z}$, but we enforce the constraint that we do not have access to analogous dataset $\mathcal{D}_{\texttt{Y} \times \texttt{Z}}$. Current state-of-the-art UNMT models divide their training procedure into two phases: \textit{i}) the \emph{pre-training} phase, in which an initial translation model is learned through a combination of language modeling or noisy reconstruction objectives \cite{song19,lewis19,lample19} applied to the monolingual data; \textit{ii}) the \emph{fine-tuning} phase, which resumes training the translation model built from the pre-training phase with a new set of objectives, typically centered around iterative back-translation i.e. penalizing a model's error in round-trip translations. We outline the objectives below:

\paragraph{Pre-training objectives} We use the \emph{MASS} objective \cite{song19}, which consists of masking\footnote{We choose a starting index of less than half the length $l$ of the input and replace the next $l/2$ tokens with a \texttt{[MASK]} token. The starting index is randomly chosen to be 0 or $l/2$ with 20\% chance for either scenario otherwise it is sampled uniformly at random.} a contiguous segment of the input and penalizing errors in the reconstruction of the masked segment. If we denote the masking operation by \texttt{MASK}, then we write the objective as follows: $$\mathcal{L}_{\text{MASS}}(x) = - \log p_{\theta}(x| \texttt{MASK}(x), l_x)$$ where $l_x$ denotes the language indicator of example $x$. We also use cross-entropy on the available auxiliary parallel data. 

\paragraph{Fine-tuning objectives} We use \emph{on-the-fly back-translation}, which we write explicitly as: $$\mathcal{L}_{\text{back-translation}}(x, l_y) = - \log p_{\theta}(x | \tilde{y}(x), l_x)$$ where  $\tilde{y}(x) = \text{argmax}_y p_{\theta}(y|x, l_y)$ and we apply a stop-gradient to $\tilde{y}(x)$. Computing the mode $\tilde{y}(x)$ of $p_{\theta}( \cdot | x, l_y)$ is intractable, so we approximate this quantity with a greedy decoding procedure. We also utilize cross-entropy, coupled with \emph{cross-translation} \cite{garcia20,li2020reference,xu19,bai2020unsupervised}, which ensures cross-lingual consistency: $$\mathcal{L}_{\text{cross-translation}} (x,y, l_z) = - \log p_{\theta}(y | \tilde{z}(x), l_y)$$ where $\tilde{z}(x) = \text{argmax}_{z} p_{\theta}(z|x, l_z)$.

\section{Method}
\label{sec:method}

For the rest of this work, we assume that we want to translate between English (\En) and some low-resource languages which we denote by \X. In our early experiments, we found that proceeding to the fine-tuning stage immediately after pre-training with MASS provided sub-optimal results (see \S\ref{subsec:synth_studies}), so we introduced an intermediate stage which leverages synthetic data to improve performance. This yields a total of three stages, which we describe below. 

\subsection{First stage of training}

In the first stage, we leverage monolingual and auxiliary parallel data, using the MASS and cross-entropy objectives on each type of dataset respectively. We describe the full procedure in Algorithm 1. \begin{algorithm}[ht]
\small
  \textbf{Input}: Datasets $\mathfrak{D}$ , number of steps $N$,  parameterized family of translation models $p_{\theta}$\; 
 \caption{ \textsc{Stage 1 \& 2}}
  \begin{algorithmic}[1]\label{algo:pretrain}
    \STATE Initialize $\theta \leftarrow \theta_0$.
    \FOR{step in 1, 2, 3, ..., $N$}
      \STATE Choose dataset $D$ at random from $\mathfrak{D}$. 
      \IF{$D$ consists of monolingual data} 
        \STATE Sample batch $x$ from $D$.
        \STATE MASS Loss: $\text{ml} \leftarrow \mathcal{L}_{\text{MASS}}(x)$.
        \STATE Update: $\theta \leftarrow \text{optimizer\_update}(\text{ml}, \theta)$.
      \ELSIF{$D$ consists of auxiliary parallel data}
        \STATE Sample batch $(x,z)$ from $D$. 
        \STATE $\text{tl} \leftarrow \mathcal{L}_{\text{cross-entropy}}(x,z) + \mathcal{L}_{\text{cross-entropy}}(z,x)$.
        \STATE Update: $\theta \leftarrow \text{optimizer\_update}(\text{tl}, \theta).$
      \ENDIF
    \ENDFOR
  \end{algorithmic}
\end{algorithm}

\begin{algorithm}[t!]
\small
 \caption{ \textsc{Stage 3} }
  \textbf{Input}: Datasets $\mathfrak{D}$, languages $\mathfrak{L}$, parameterized family of translation models $p_{\theta}$, initial parameters from pre-training $\theta_0$ \; 
  \begin{algorithmic}[1] \label{algo:finetune}
    \STATE Initialize $\theta \leftarrow \theta_0$.
    \STATE Target Languages: $\mathfrak{L}_T \leftarrow \{ \text{Gu, Kk, Ne, Si, Tr} \}.$
    \WHILE{not converged} 
      \FOR{$D$ in $\mathfrak{D}$}
          \IF{$D$ consists of monolingual data} 
            \STATE  $l_{D} \leftarrow $ Language of $D$.
            \STATE  Sample batch $x$ from $D$.
            \IF{$l_{D}$ is English}
              \FOR{$l$ in $\mathfrak{L}_{T}, l \neq l_{D}$}
                \STATE Translation: $\hat{y}_l \leftarrow $Decode $p_{\theta}( \hat{y}_l | x).$
                \STATE $\text{bt} \leftarrow \mathcal{L}_{\text{back-translation}}(x,l).$
                \STATE Update: $\theta \leftarrow \textnormal{optimizer\_update}(\text{bt}, \theta).$
              \ENDFOR
            \ELSE
               \STATE $\mathfrak{R}_{D} \leftarrow \text{Auxiliary languages for } l_D.$
                \FOR{$l$ in $\mathfrak{R}_{D} \cup \text{English}$}
                  \STATE Translation: $\hat{y}_l \leftarrow $Decode $p_{\theta}( \hat{y}_l | x).$
                  \STATE $\text{bt} \leftarrow \mathcal{L}_{\text{back-translation}}(x,l).$
                  \STATE Update: $\theta \leftarrow \textnormal{optimizer\_update}(\text{bt}, \theta).$
                \ENDFOR
            \ENDIF
          \ELSIF{$D$ consists of parallel data}
            \STATE Sample batch $(x,z)$ from $D$.
            \STATE Source language: $l_x \leftarrow$ Language of $x$.
            \STATE Target language: $l_z \leftarrow$ Language of $z$.
            \IF{$D$ is not synthetic}
              \FOR{$l$ in $\mathfrak{L}, l \neq l_x, l_z$}
                \STATE $\text{ct} \leftarrow \mathcal{L}_{\text{cross-translation}}(x,z,l).$
                \STATE Update: $\theta \leftarrow \textnormal{optimizer\_update}(\text{ct}, \theta).$
              \ENDFOR
            \ELSE
              \STATE Cross-entropy: $\text{ce} \leftarrow \mathcal{L}_{\text{cross-entropy}}(x,z).$
              \STATE Update: $\theta \leftarrow \textnormal{optimizer\_update}(\text{ce}, \theta).$
            \ENDIF
          \ENDIF
        \ENDFOR
    \ENDWHILE
  \end{algorithmic}
\end{algorithm}
\subsection{Second stage of training}
Once we have completed the first stage, we will have produced an initial model capable of generating high-quality \X $\rightarrow$\En \ (to English) translations for all of the low-resource pairs we consider, also known as many-to-one setup in multilingual NMT \cite{johnson16}. Unfortunately, the model does not reach that level of performance for the \En $\rightarrow$\X \, translation directions, generating very low-quality translations into these low-resource languages. Note that, this phenomenon is ubiquitously observed in multilingual models \cite{firat16,johnson16,aharoni19}. This abysmal performance could have dire consequences in the fine-tuning stage, since both on-the-fly back-translation and cross-translation rely heavily on intermediate translations. We verify that this is in fact the case in \S\ref{subsec:synth_studies}.

Instead, we exploit the strong \X $\rightarrow$\En \,performance by translating subsets\footnote{We utilize 10\% of the monolingual data for each low-resource language.} of the monolingual data of the low-resource languages using our initial model and treat the result as pseudo-parallel datasets for the language pairs \En $\rightarrow$\X. More explicitly, given a sentence $x$ from a low-resource language, we generate an English translation $\tilde{y}_{\texttt{En}}$ with our initial model and create a synthetic translation-pair $(\tilde{y}_{\texttt{En}}, x)$. We refer to this procedure as \emph{offline back-translation} \cite{sennrich2015improving}. We add these datasets to our collection of auxiliary parallel corpora and repeat the training procedure from the first stage (Algorithm 1), starting from the last checkpoint. Note that, while  offline back-translated (synthetic) data is commonly  used for zero-resource translation \cite{firat-etal-2016-zero,DBLP:journals/corr/ChenLCL17}, it is worth emphasizing the difference here again, that in the configuration studied in this paper, we do not assume the existence of any parallel data between \En $\leftrightarrow$\X, which is exploited by such methods. 

Upon completion, we run the procedure a second time, with a new subset of synthetic data of twice the size for the \En $\rightarrow$\X \, pairs. Furthermore, since the translations from English have improved, we take disjoint subsets\footnote{1 million lines of English per low-resource language.} of the English monolingual data and generate corpora of synthetic \X $\rightarrow$\En \, translation pairs that we also include in the second run of our procedure. 

\subsection{Third stage of training}

For the third and final stage of training, we use back-translation of the monolingual data and cross-translation\footnote{For Nepali, Sinhala and Gujarati, we use Hindi as the pivot language. For Turkish, we use Arabic and for Kazakh, we use Russian.} on the auxiliary parallel data. We also leverage the synthetic data through the cross-entropy objective. We present the procedure in detail under Algorithm 2.

\section{Main experiment}

In this section, we describe the details of our main experiment. As indicated in Figure~\ref{fig:langgraph}, we consider five languages (Nepali, Sinhala, Gujarati, Kazakh, Turkish) as the target unsupervised language pairs with English. We leverage auxiliary parallel data from six higher-resource languages (Chinese, Russian, Arabic, Hindi, Telugu, Tamil) with English. The domains and counts for the datasets considered can be found in Table \ref{tab:languages} and a more detailed discussion on the source of the data and the preprocessing steps can be found in the Appendix.  In the following subsections, we provide detailed descriptions of the model configurations, training parameters, evaluation and discuss results of our main experiment.

\begin{table*}
\small
\centering
\begin{tabular}{llccccccc}
\toprule
 & \textbf{Model} & \multicolumn{2}{l}{\stackanchor{\, \emph{newstest2019}}{\Gu \, $\leftrightarrow$ \En}} &  \multicolumn{2}{l}{\, \stackanchor{\emph{newstest2019}}{\Kk \, $\leftrightarrow$ \En}} &  \multicolumn{2}{l}{\, \stackanchor{\emph{newstest2017}}{\Tr \, $\leftrightarrow$ \En}} \\ \midrule
\multirow{1}{*}{No parallel data}  & \citet{kim2020and} & 0.6 & 0.6 & 0.8 & 2.0 & - & -  \\ \hline
\multirow{3}{*}{\stackanchor{No parallel data}{for\{\Gu,\Kk,\Tr\}}}  &  \emph{Stage 1} (Ours)  & 4.4 & 19.3 & 3.9 & 14.8 & 8.4  & 15.9 \\
& \emph{Stage 2} (Ours)& \textbf{16.4} & 20.4 & 9.9 & 15.6 & \textbf{20.0} & \textbf{20.5}  \\
 & \emph{Stage 3} (Ours) & \textbf{16.4} & \textbf{22.2} & \textbf{10.4} & \textbf{16.4} & 19.8 & 19.9  \\ \hline
\multirow{2}{*}{\stackanchor{With parallel data}{for\{\Gu,\Kk,\Tr\}}} 
& \emph{Mult. MT Baseline} (Ours) & \underline{15.5} & \underline{19.3} & \underline{9.5} & \underline{15.1} &  \underline{18.1} & 22.0  \\
& mBART \cite{liu2020multilingual}  &  0.1 & 0.3 & 2.5 & 7.4 &  17.8 & \underline{22.5}  \\  \bottomrule
\end{tabular}
\caption{BLEU scores of various supervised and unsupervised models on the WMT \emph{newstest} sets. The bolded numbers are the best unsupervised scores and the underlined numbers represent the best supervised scores. For any \X $\leftrightarrow$\Y \, language pair, the \X $\rightarrow$\Y \, translation results are listed under each \Y \, column, and vice-versa.}
\label{tab:benchmark-bleu-wmt}
\end{table*}

\begin{table*}[!htbp]
\centering
\small
\begin{tabular}{llcccc}
\toprule
 & \textbf{Model} &  \multicolumn{2}{l}{\stackanchor{\emph{FLoRes devtest}}{\Ne \, $\leftrightarrow$ \En}} &  \multicolumn{2}{l}{\stackanchor{\emph{FLoRes devtest}}{\Si \, $\leftrightarrow$ \En}}\\ \midrule
\multirow{1}{*}{No parallel data} & \citet{guzman2019flores}  & 0.1 & 0.5 & 0.1 & 0.1 \\  \hline
\multirow{5}{*}{\stackanchor{No parallel data}{with\{\Ne,\Si\}}}&\citet{liu2020multilingual} &  - & 17.9 & - & 9.0 \\ 
 & \citet{guzman2019flores} & 8.3 & 18.3 & 0.1 & 0.1 \\  
& \emph{Stage 1} (Ours) & 3.3 & 18.3 & 1.4 & 11.5 \\
& \emph{Stage 2} (Ours)& 8.6  & 20.8 & 7.7 & 15.7  \\
 & \emph{Stage 3} (Ours) &  \textbf{8.9} & \textbf{21.7} & \textbf{7.9} & \textbf{16.2}  \\ \hline
\multirow{3}{*}{\stackanchor{With parallel data}{for\{\Ne,\Si\}}} & \emph{Mult. MT Baseline} (Ours) &  8.6 & 20.1 & 7.6 & 15.3 \\
& \citet{liu2020multilingual} & \underline{9.6} & 21.3 & \underline{9.3} & \underline{20.2} \\
& \citet{guzman2019flores} & 8.8 & \underline{21.5} & 6.5 & 15.1 \\ \bottomrule
\end{tabular}
\caption{BLEU scores of various supervised and unsupervised models on the FLoRes \emph{devtest} sets. The bolded numbers are the best unsupervised scores and the underlined numbers represent the best supervised scores. For any \X $\leftrightarrow$\Y \, language pair, the \X $\rightarrow$\Y \, translation results are listed under each \Y \, column, and vice-versa.}
\label{tab:benchmark-bleu-flores}
\end{table*}

\subsection{Datasets and preprocessing}

We draw most of our data from WMT. The monolingual data comes from News Crawl\footnote{http://data.statmt.org/news-crawl/} when available. For all the unsupervised pairs except Turkish, we supplement the News Crawl datasets with monolingual data from Common Crawl and Wikipedia\footnote{We used the monolingual data available from https://github.com/facebookresearch/flores for Nepali and Sinhala in order to avoid any data leakage from the test sets.}. The parallel data we use came from a variety of sources, all available through WMT. We drew our English-Hindi parallel data from IITB \cite{kunchukuttan2017iit}; English-Russian, English-Arabic, and English-Chinese parallel data from the UN Corpus \cite{ziemski16}; English-Tamil and English-Telugu from Wikimatrix \cite{schwenk2019wikimatrix}. We used the scripts from Moses \cite{koehn09} to normalize punctuation, remove non-printing characters, and replace the unicode characters with their non-unicode equivalent. We additionally use the normalizing script from Indic NLP \cite{kunchukuttan2020indicnlp} for Gujarati, Nepali, Telugu, and Sinhala. 

We concatenate two million lines of monolingual data for each language and use it to build a vocabulary with SentencePiece\footnote{We build the SentencePiece model with the following settings: vocab\_size=64000, model\_type=bpe, user\_defined\_symbols=[MASK], character\_coverage=1.0, split\_by\_whitespace=true.} \cite{kudo18} of 64,000 pieces. We then separate our data into SentencePiece pieces and remove all training samples that are over 88 pieces long. 

\subsection{Model architecture}
All of our models were coded and tested in Tensorflow \cite{abadi2016tensorflow}. We use the Transformer architecture \cite{vaswani17} as the basis of our translation models. We use 6-layer encoder and decoder architecture with a hidden size of 1024 and an 8192 feedforward filter size. We share the same encoder for all languages. To differentiate between the different possible output languages, we add (learned) language embeddings to each token's embedding before passing them to the decoder. We follow the same modification as done in~\citet{song19} and modify the output transformation of each attention head in each transformer block in the decoder to be distinct for each language. Besides these modifications, we share decoder parameters for every language. 

\subsection{Training parameters}

We use three different settings, corresponding to each stage of training. For the first stage, we use the Adam optimizer \cite{kingma14} with a learning rate of 0.0002, weight decay of 0.2 and batch size of 2048 examples. We use a learning rate schedule consisting of a linear warmup of 4000 steps to a value 0.0002 followed by a linear decay for 1.2 million steps. At every step, we choose a single dataset from which to draw a whole batch using the following process: with equal probability, choose either monolingual or parallel. If the choice is monolingual, then we select one of the monolingual datasets uniformly at random. If the choice is parallel, we use a temperature-based sampling scheme based on the numbers of samples with a temperature of 5 \cite{arivazhagan19a}. In the second stage, we retain the same settings for both rounds of leveraging synthetic data except for the learning rate and number of steps. In the first round, we use the same number of steps, while in the second round we only use 240 thousand steps, a 1/5th of the original. 

For the final phase, we bucket sequences by their sequence length and group them up into batches of at most 2000 tokens. We train the model with 8 NVIDIA V100 GPUs, assigning a batch to each one of them and training synchronously. We also use the Adamax optimizer instead, and cut the learning rate by four once more. 

\subsection{Baselines}

We compare with the state-of-the-art unsupervised and supervised baselines from the literature. Note all the baselines build language-specific models, whereas we have a single model for all the target unsupervised directions. 

\paragraph{Unsupervised baselines:} For the bilingual unsupervised baselines, we include the results of \citet{kim2020and}\footnote{Due to the limited literature on unsupervised machine translation on low-resource languages, this was the best bilingual unsupervised system we could find.} for \En $\leftrightarrow$\Gu \ and \En $\leftrightarrow$\Kk \, and of \citet{guzman2019flores} for \En $\leftrightarrow$\Si. We also report other multilingual unsupervised baselines. mBART \cite{liu2020multilingual} leverages auxiliary parallel data (e.g. \En $\leftrightarrow$\Hi \, parallel data for \Gu $\rightarrow$\En) after pre-training on a large dataset consisting of 25 languages and the FLoRes dataset benchmark~\cite{guzman2019flores} leverages \Hi $\leftrightarrow$\En \, data for the \En $\leftrightarrow$\Ne \ language pair. All the unsupervised baselines that use auxiliary parallel data perform considerably better than the ones that don't.  

\paragraph{Supervised baselines:} In addition to the unsupervised numbers above, mBART and the FLoRes dataset benchmarks report supervised results that we compare with. We additionally include one more baseline where we followed the training scheme proposed in stage 1, but also included the missing parallel data. We labeled this model ``Mult. MT Baseline", though we emphasize that we also leverage the monolingual data in this baseline, as in recent work \cite{siddhant-etal-2020-leveraging,garcia20}.

\subsection{Evaluation}
We evaluate the performance of our models using BLEU scores \cite{papineni2002bleu}. BLEU scores are known to be dependent on the data pre-processing \cite{post-2018-call} and thus proper care is required to ensure the scores between our models and the baselines are comparable. We thus only considered baselines which report detokenized BLEU scores with sacreBLEU \cite{post-2018-call} or report explicit pre-processing steps. In the case of the Indic languages (Gujarati, Nepali, and Sinhala), both the baselines we consider \cite{guzman2019flores,liu2020multilingual} report tokenized BLEU using the tokenizer provided by the Indic-NLP library \cite{kunchukuttan2020indicnlp}. For these languages, we follow this convention as well so that the BLEU scores remain comparable. Otherwise, we follow suit with the rest of the literature and report detoknized BLEU scores through sacreBLEU\footnote{BLEU + case.mixed + numrefs.1 + smooth.exp + tok.13a + version.1.4.14}.

\subsection{Results \& discussion}

We list the results of our experiments for the WMT datasets in Table \ref{tab:benchmark-bleu-wmt} and for the FLoRes datasets in Table \ref{tab:benchmark-bleu-flores}. After the first stage of training, we obtain competitive BLEU scores for \X $\rightarrow$\En \, translation directions, outperforming all unsupervised models as well as mBART for the language pairs \Kk $\rightarrow$\En \, and \Gu$ \rightarrow$\En. Upon completion of the second stage of training, we see that the \En $\rightarrow$\X \, language pairs observe large gains, while the \X $\rightarrow$\En \ directions also improve. The final round of training further improves results in some language pairs, yielding an increase of +0.44 BLEU on average. 

Note that in addition to considerably outperforming all the unsupervised baselines, our approach outperforms the supervised baselines on many of the language pairs, even matching the state-of-the-art on \Ne $\rightarrow$\En. Specifically, it outperforms the supervised mBART on six out of ten translation directions despite being a smaller model and \citet{guzman2019flores} on all pairs. Critically, we outperform our own multilingual MT baseline, trained in the same fashion and data as Stage 1, which further reinforces our assertion that unsupervised MT can provide competitive results with supervised MT in low-resource settings.


\begin{table}
\small
\centering
\begin{tabular}{llll}
\toprule
\multicolumn{2}{l}{\stackanchor{\textbf{Data configuration}}{Monolingual \quad Parallel}} & \multicolumn{2}{l}{\stackanchor{\emph{newsdev2019}}{\Kk$\leftrightarrow$\En}}  \\ \midrule 
\Ru & \Ru & 6.8 & 9.5 \\ 
\Ru, \Ar, \Zh & \Ru & 7.3 & 14.8 \\
\Ru & \Ru, \Ar, \Zh & 9.6 & 18.4 \\
\Ru, \Ar, \Zh  & \Ru, \Ar, \Zh   & 9.8 & 18.6 \\ \bottomrule
\end{tabular}
  \caption{BLEU scores for a model trained with various configurations for the auxiliary data.}
  \label{tab:multilingual_effects}
\end{table}

\section{Further analysis}

\begin{table*}[!htbp]
\centering
\small
\begin{tabular}{llcccccccccc}
\toprule
 & \textbf{Stage} & \multicolumn{2}{l}{\stackanchor{\, \, \emph{newsdev2019}}{\Gu \, $\leftrightarrow$ \En}} &  \multicolumn{2}{l}{\stackanchor{\, \emph{newsdev2019}}{\, \Kk \, $\leftrightarrow$ \En}} &  \multicolumn{2}{l}{\, \stackanchor{\emph{newsdev2016}}{\Tr \, $\leftrightarrow$ \En}} & \multicolumn{2}{l}{\stackanchor{\,  \emph{FLoRes dev}}{\, \Ne \, $\leftrightarrow$ \En}} & \multicolumn{2}{l}{\, \stackanchor{\emph{FLoRes dev}}{\Si \, $\leftrightarrow$ \En}}\\ \midrule
\multirow{1}{*}{Baseline} & First  & 5.0 & 23.4 & 4.0 & 16.1 & 6.3  & 17.7 & 2.8 & 15.1 & 1.3 & 12.0  \\ \hline
\multirow{2}{*}{Without synthetic data} & Second  & 6.2 & 24.8 & 4.24 & 17.0 & 6.3 & 18.5 & 3.6 & 16.0 & 1.6 & 12.7 \\
& Third  & 12.9 & 26.2 &  6.1 &  16.3 &  12.9 & 19.5 & 5.9 & 16.1 & 5.2 & 13.1 \\ \hline
\multirow{2}{*}{With synthetic data} & Second & 19.6 & 29.8 & 10.6 & 20.0 & 16.7 & 23.8 & 7.3 & 17.4 & 8.3 & 16.6 \\
& Third & 18.6 & 30.3 & 11.6 & 21.5 & 17.9 & 24.7 & 8.2 & 17.6  & 7.7 & 17.4 \\ \bottomrule

\end{tabular}
\caption{BLEU scores of model configurations with or without synthetic data. Otherwise, we report the numbers after stage 2 for both models and use the results after stage 1 as a baseline.}
\label{tab:synth_studies}
\end{table*}

Given the substantial quality gains delivered by our proposed method, we set out to investigate what design choices can improve the performance of unsupervised models. To ease the computational burden, we further filter the training data to remove any sample which are longer than 64 SentencePiece\footnote{For all the experiments in this section, we use the same SentencePiece vocabulary as our benchmark model.} pieces long and cut the batch size in half for the first two stages. Additionally, we only do one additional round of training with synthetic data as opposed to the two rounds performed for the benchmark models. While these choices negatively impact performance, the resulting models still provide competitive results with our baselines and hence are more than sufficient for the purposes of experimental studies.

\subsection{Increasing multilinguality of the auxiliary parallel data improves performance}

It was shown in \citet{garcia20, bai2020unsupervised} that adding more multilingual data improved performance, and that the inclusion of auxiliary parallel data further improved the BLEU scores \cite{siddhant2020leveraging}. In this experiment, we examine whether further increasing multilinguality under a fixed data budget improves performance. For all configurations in this subsection, we utilize all the available English and Kazakh monolingual data. We fix the amount of auxiliary monolingual data to 40 million, the auxiliary parallel data to 12 million, and vary the number of languages which manifest in this auxiliary data.


We report the results on Table \ref{tab:multilingual_effects}. It is observed that increasing the multilinguality of the parallel data is crucial, but the matter is less clear for the monolingual data. Using more languages for the monolingual data can potentially harm performance, but in the presence of multiple auxiliary language pairs with supervised data this degradation vanishes.

\subsection{Synthetic data is critical for both stage 2 and stage 3 of training}\label{subsec:synth_studies}
In the following experiments, we evaluate the role of synthetic parallel data in the improved performance found at the end of stage 2 and stage 3 of our training procedure. We first evaluate whether the improved performance at the end of stage 2 comes from the synthetic data or the continued training. We consider the alternative where we repeat the same training steps as in stage 2 but without the synthetic data. We then additionally fine-tune these models with the same procedure as stage 3, but without any of the terms involving synthetic data. We report the BLEU scores for all these configurations in Table \ref{tab:synth_studies}. The results suggest: the baseline without synthetic parallel data shows inferior performance across all language pairs compared to our approach leveraging synthetic parallel data.

Finally, we inspect whether the synthetic parallel data is still necessary in stage 3 or if it suffices to only leverage it during the second stage. We consider three fine-tuning strategies, where we either \textbf{(1)} only utilize on-the-fly back-translation \textbf{(2)} additionally include cross-translation terms for Gujarati, Nepali, and Sinhala using Hindi \textbf{(3)} additionally include a cross-translation terms for Turkish and Kazakh involving Arabic and Russian respectively. We compare all of the approaches to the vanilla strategy that only leverages on-the-fly back-translation and report the aggregate improvements in BLEU on the \X $\rightarrow$\En \, directions over this baseline in Table \ref{tab:finetuning_strats}. We see two trends: The configurations that do not leverage synthetic data perform worse than those that do, and increasing multilinguality through the inclusion of cross-translation further improves performance.

\subsection{Our approach is robust under multiple domains}
\begin{table}
\small
\centering
\begin{tabular}{llll}
\toprule
& \textbf{Objectives} & \multicolumn{1}{l}{$\Delta$\text{BLEU}} \\ \midrule
 \multirow{3}{*}{\stackanchor{With}{synthetic data}} & BT & 0.6 \\ 
 & + CT with \Hi &  2.1  \\ 
 & + CT with \Ru \,and \Ar & 2.6 \\ \hline
 \multirow{3}{*}{\stackanchor{Without}{synthetic data}} & BT & 0.0 \\ 
 & + CT with \Hi &  1.3  \\ 
 & + CT with \Ru \,and \Ar & 1.4 \\ \bottomrule
\end{tabular}
  \caption{Total BLEU increase for \X $\rightarrow$\En \, over baseline fine-tuning strategy consisting of on-the-fly back-translation (BT) and no synthetic data. We refer to cross-translation as "CT".}
  \label{tab:finetuning_strats}
\end{table}

\begin{table}
\centering
\small
\begin{tabular}{llllll}
\toprule
\textbf{Data Configurations} & \multicolumn{2}{l}{\stackanchor{\emph{newstest2019}}{\Gu $\leftrightarrow$ \En}} & \multicolumn{2}{l}{\stackanchor{\emph{newsdev2019}}{\Gu $\leftrightarrow$ \En}}  \\ \midrule
500k News Crawl & 6.8 & 15.7 & 9.7 & 21.7 \\
500k Common Crawl & 9.2 & 16.7 & 9.4 & 22.5 \\
100k News Crawl & 3.6 & 10.0 & 5.4 & 12.4 \\ \hline
mBART  & \quad - & 13.8 & \quad - & \quad - \\
\citet{kim2020and} & 0.6 & 0.6  & \quad - & \quad -  \\ \bottomrule
\end{tabular}
  \caption{BLEU scores for various configurations of Gujarati monolingual data, where we vary amount of data and domain. We include the best results of mBART and \cite{kim2020and} for comparison.  }
  \label{tab:domain_effects}
\end{table}

We investigate the impact of data quantity and quality on the performance of our models. In this experiment, we focus on \En $\leftrightarrow$\Gu \ and use all available monolingual and auxiliary parallel data for all languages except Gujarati. We consider three configurations: \textbf{(1)} 500,000 lines from News Crawl (in-domain high-quality data); \textbf{(2)} 500,000 lines from Common Crawl (multi-domain data); \textbf{(3)} 100,000 lines from News Crawl. We present the results on both \emph{newstest2019} and \emph{newsdev2019} for \En $\leftrightarrow$\Gu \, on Table \ref{tab:domain_effects}. We see that both Common Crawl and News Crawl configurations produce similar results at this scale, with the Common Crawl configuration having a small edge on average. Notice that even in this data-starved setting, we still outperform the competing unsupervised models. Once we reach only 100,000 lines, performance degrades below mBART but still outperforms the bilingual UNMT approach of \citet{kim2020and}, revealing the power of multilinguality in low-resource settings.

\section{Conclusion}

In this work, we studied how multilinguality can make unsupervised translation viable for low-resource languages in a realistic setting. Our results show that utilizing the auxiliary parallel data in combination with synthetic data through our three-stage training procedure not only yields large gains over unsupervised baselines but also outperforms several modern supervised approaches.

\bibliography{anthology,custom}
\bibliographystyle{acl_natbib}
\end{document}